\PassOptionsToPackage{table,xcdraw}{xcolor}
\documentclass[sigconf,screen,authorversion]{acmart}

\AtBeginDocument{%
  }

\copyrightyear{2026}
\acmYear{2026}
\setcopyright{cc}
\setcctype{by}
\acmConference[LAK 2026]{LAK26: 16th International Learning Analytics and Knowledge Conference}{April 27-May 01, 2026}{Bergen, Norway}
\acmBooktitle{LAK26: 16th International Learning Analytics and Knowledge Conference (LAK 2026), April 27-May 01, 2026, Bergen, Norway}
\acmPrice{}
\acmDOI{10.1145/3785022.3785083}
\acmISBN{979-8-4007-2066-6/2026/04}

\begin{document}

\title{Using Large Language Models to Detect Socially Shared Regulation of Collaborative Learning}

\author{Jiayi Zhang}
\orcid{0000-0002-7334-4256}
\affiliation{
  \institution{University of Pennsylvania}
  \city{Philadelphia, PA}
  \country{USA}
}
\email{joycez@upenn.edu}

\author{Conrad Borchers}
\orcid{0000-0003-3437-8979}
\affiliation{
  \institution{Carnegie Mellon University}
  \city{Pittsburgh, PA}
  \country{USA}
}
\email{cborcher@cs.cmu.edu}

\author{Clayton Cohn}
\orcid{0000-0003-0856-9587}
\affiliation{
  \institution{Vanderbilt University}
  \city{Nashville, TN}
  \country{USA}
}
\email{clayton.a.cohn@vanderbilt.edu}

\author{Namrata Srivastava}
\orcid{0000-0003-4194-318X}
\affiliation{
  \institution{Vanderbilt University}
  \city{Nashville, TN}
  \country{USA}
}
\email{Namrata.Srivastava@vanderbilt.edu}

\author{Caitlin Snyder}
\orcid{0000-0002-3341-0490}
\affiliation{
  \institution{University of Detroit Mercy}
  \city{Detroit, MI}
  \country{USA}
}
\email{snydercr@udmercy.edu}

\author{Siyuan Guo}
\orcid{0009-0001-4305-9147} 
\affiliation{
  \institution{Vanderbilt University}
  \city{Nashville, TN}
  \country{USA}
}
\email{siyuan.guo@vanderbilt.edu}

\author{Ashwin T S}
\orcid{0000-0002-1690-1626}
\affiliation{
  \institution{Vanderbilt University}
  \city{Nashville, TN}
  \country{USA}
}
\email{ashwindixit9@gmail.com}

\author{Naveeduddin Mohammed}
\orcid{0000-0002-3706-2884}
\affiliation{
  \institution{Vanderbilt University}
  \city{Nashville, TN}
  \country{USA}
}
\email{naveeduddin.mohammed@vanderbilt.edu}

\author{Haley Noh}
\orcid{0009-0004-1092-3041}
\affiliation{
  \institution{New York University}
  \city{New York, NY}
  \country{USA}
}
\email{hhn2020@nyu.edu}

\author{Gautam Biswas}
\orcid{0000-0002-2752-3878}
\affiliation{
  \institution{Vanderbilt University}
  \city{Nashville, TN}
  \country{USA}
}
\email{gautam.biswas@vanderbilt.edu}

\makeatletter
\let\@authorsaddresses\@empty
\makeatother

\renewcommand{\shortauthors}{Zhang et al.}

\begin{abstract}
The field of learning analytics has made notable strides in automating the detection of complex learning processes in multimodal data. However, most advancements have focused on individualized problem-solving instead of collaborative, open-ended problem-solving, which may offer both affordances (richer data) and challenges (low cohesion) to behavioral prediction. Here, we extend predictive models to automatically detect socially shared regulation of learning (SSRL) behaviors in collaborative computational modeling environments using embedding-based approaches. We leverage large language models (LLMs) as summarization tools to generate task-aware representations of student dialogue aligned with system logs. These summaries, combined with text-only embeddings, context-enriched embeddings, and log-derived features, were used to train predictive models. Results show that text-only embeddings often achieve stronger performance in detecting SSRL behaviors related to enactment or group dynamics (e.g., off-task behavior or requesting assistance). In contrast, contextual and multimodal features provide complementary benefits for constructs such as planning and reflection. %
Overall, our findings highlight the promise of embedding-based models for extending learning analytics by enabling scalable detection of SSRL behaviors, ultimately supporting real-time feedback and adaptive scaffolding in collaborative learning environments that teachers value.
\end{abstract}



\maketitle

\section{Introduction}
\label{sec:intro}

A core goal of learning analytics is to understand complex processes that either inhibit or foster learning. Our field is increasingly turning to complex multimodal data and learning systems to generate such insights. These data afford complex methods for modeling and prediction. As an emerging example, collaborative computational modeling, where students work together to model and analyze scientific processes, has been shown to foster Science, Technology, Engineering, Mathematics, and Computing (STEM+C) learning \cite{snyder2024analyzing}.

While prior work has demonstrated that open-ended problem-solving environments often improve students’ understanding of science and computing \cite{hutchins2020domain}, these environments also introduce additional challenges and complexities. In these environments, students need to integrate domain-specific scientific representations with domain-general computational concepts and coordinate multiple epistemic practices (such as model design, simulation, verification, and revision) within a single learning activity \cite{basu2016identifying}.

To better understand how students learn in complex STEM+C settings, it is essential to capture the processes and behaviors that shape their learning. Collaborative discourse provides a powerful lens for this work, offering rich data through student conversations that reveal metacognitive behaviors, such as planning, monitoring, and reflecting. These insights can help elucidate how students learn in authentic collaborative contexts, enabling the development of supports that address their needs and promote deeper integration of conceptual knowledge \cite{huang2025examining,hadwin2017self,panadero2015socially}.

However, discourse alone is often insufficient because students may struggle to express their thoughts, and their verbalizations frequently lack context related to the specific problem-solving activity in the learning environment. For instance, in the context of computational modeling, a student saying ``put that block there'' lacks explicit reference to the specific computational model components and domain constructs being discussed. 

Integrating multiple modalities to inform and contextualize insights through multimodal learning analytics (MMLA) offers a promising solution, as previous work has shown that combining collaborative discourse with environment logs enables the grounding of discussions in verifiable data from students' actions within the learning environment \cite{cohn2024towards}. While MMLA has been extensively used to gain insights into student learning and to understand their self-regulated learning (SRL) processes \cite{Zimmerman2001}, few researchers have fully leveraged MMLA to explore the rich opportunities provided by collaborative discourse in STEM+C contexts. 

In this paper, we leverage large language models to explore how MMLA can be used to train models that detect students’ socially shared regulation of learning (SSRL) behaviors by integrating discourse with log data from a K-12 C2STEM collaborative computational modeling environment \cite{hutchins2020c2stem}. We focus on SSRL processes that emerge in collaboration, including \emph{planning}, \emph{enacting}, \emph{monitoring}, and \emph{reflecting}, as well as \emph{assistance} and \emph{off-topic} talk, which shape group dynamics. Our goal is to demonstrate the feasibility of embedding-based models for detecting SSRL behaviors at scale in collaborative learning environments. Such models can extend MMLA by examining how groups regulate their work toward shared goals and by supporting timely feedback and interventions. To enhance the adaptability of our models, we conducted interviews with teachers to gather feedback on their perceptions of the usability of these models.

\section{Background}
\label{sec:background}

\subsection{Self-Regulated Learning and Socially Shared Regulation of Learning}

\textbf{Self-Regulated Learning}. In individual learning contexts, self-regulated learning refers to a student’s ability to actively monitor and regulate their attention and effort in pursuit of a goal. During this process, learners may set goals and make plans, enact those plans, and reflect and adjust when goals are not met \cite{Zimmerman2001}. Prior research has consistently demonstrated that students who engage in these SRL strategies tend to exhibit improved learning across diverse contexts \cite{Zimmerman2001, VanderGraaf2022, Heirweg2020}. However, SRL enactment varies widely across learners, and many students struggle to consistently apply effective regulatory strategies \cite{greene2017capturing}.

In response, the field of learning analytics has explored methods to measure and scaffold SRL in a timely, adaptive fashion to facilitate self-regulatory behaviors and improve learning outcomes \cite{lim2024students, Li2024}. Several studies have developed models to infer SRL using trace data and students' utterances from think-aloud protocols \cite{Winne2000, Azevedo2010}. For example, researchers used behavioral features (response timing and hint usage) to automatically detect planning and reflection \cite{Winne2000,lim2024students}. More recently, \cite{zhang2024using} examined the potential to automatically infer SRL from students' think-aloud utterances using machine-learning models. Together, these computational approaches offer a promising pathway toward scalable, real-time SRL detection, enabling adaptive support and personalized interventions.

\textbf{Socially Shared Regulation of Learning}. While SRL concerns how individuals regulate their learning, socially shared regulation of learning refers to regulatory processes co-constructed within collaborative contexts. SSRL occurs when group members collectively set goals, monitor progress, manage motivation, and reflect on outcomes in pursuit of a shared task \cite{hadwin2017self}. Foundational frameworks \cite{hadwin2017self} identify core SSRL categories, such as planning (negotiating goals and strategies), monitoring (evaluating progress), enacting/control (coordinating actions and maintaining focus), and reflection (reviewing outcomes). These processes unfold dynamically during group interaction and are situated in broader social and task contexts. Research shows that groups engaging in high-quality SSRL (e.g., shared goal-setting and co-monitoring) achieve deeper learning, stronger knowledge integration, and more equitable participation \cite{panadero2015socially}. %

To examine SSRL behaviors, prior work has analyzed discourse and interaction sequences, operationalizing constructs such as shared goal-setting and collective reflection through qualitative coding and sequence analysis \cite{panadero2015socially}. Yet computational modeling approaches remain limited, despite their potential to provide scalable and timely measurements. In this work, we aim to develop models that automatically detect SSRL behaviors in collaborative learning environments using multimodal data from both log traces and student dialogue, enabling real-time identification of group dynamics to inform adaptive support.

\subsection{Large Language Models for Self-Regulated Learning Measurements}
Large language models (LLMs) have recently gained attention in learning analytics for their potential to detect SRL processes. Early studies show that LLMs can interpret think-aloud protocols to identify SRL behaviors, such as planning, monitoring, and reflecting \cite{zhang2024using,borchers2024using}. 

Despite these promising developments, most applications of LLMs for SRL detection have focused on \emph{individual learning contexts}. In contrast, relatively little work has examined the role of LLMs in \emph{collaborative, multimodal environments} where self-regulation is intertwined with socially shared regulation of learning \cite{liu2025multimodal,ma2022detecting}. In such settings, students not only plan and monitor individually but also coordinate strategies, enact solutions, and reflect collectively \cite{hadwin2017self}. These behaviors are expressed through both collaborative discourse and digital traces of environmental use. However, existing LLM-based approaches rarely incorporate action logs or contextualize utterances with the specific task components being manipulated. This gap is particularly salient in \emph{computational STEM+C modeling environments}, where students engage in open-ended code- and model-building tasks that require integrating domain knowledge across science and computing. While some exploratory work has examined machine learning-based interaction analysis in \emph{embodied} learning environments \cite{fonteles2024}, LLM-based SRL detection in collaborative computational modeling settings remains less explored.

These gaps point to the need for scalable alternatives that retain the representational richness of text while integrating multimodal context from the learning environment. Our work addresses this need by using LLMs not only as the predictor but as a controlled summarization tool: we prompt an LLM to generate concise, task-aware summaries of students’ collaborative segments (aligned with their environment actions), and then use SSRL categories inferred from these summaries (together with discourse and log-derived features) to train predictive models. We conducted a systematic comparison of text-only (semantic) embeddings, context-enriched embeddings, and multimodal combinations to evaluate their effectiveness for SSRL detection.

A related limitation is that LLM-based methods, although semantically powerful, often act as black boxes, producing hallucinations and unstable predictions \cite {doshi2017towards}. These limitations hinder classroom adoption and raise concerns about usability for teachers. To explore usability, we contacted teachers, explained how predictions are generated, and gathered feedback to inform future refinements.

\section{Methods}
\label{sec:methods}

\subsection{Learning Environment and Study Design}
C2STEM \cite{hutchins2020c2stem} is a collaborative computational modeling environment designed to integrate high school physics and computing, challenging students to translate their understanding of one- and two-dimensional kinematics into computational form by using block-based code to model the motion of various objects (e.g., a truck or drone; see \ref{fig:c2stem}). 
Students are required to integrate kinematics concepts such as acceleration, position, and velocity with computing constructs like initializing variables, conditional statements, and loops. Block names like \textit{set x position to} and \textit{update velocity by} support this integration by making explicit connections between physics and computing concepts. 

This work focuses on C2STEM's \textit{Truck Task}, where students use their knowledge of one-dimensional kinematics to collaboratively (in dyads) model a truck starting from rest, accelerating to the speed limit, cruising at the speed limit for as long as possible, then decelerating to stop at a stop sign. We analyze C2STEM data collected during two studies (2021 and 2022) involving a total of 36 high school sophomores who participated in either an eight-week (2021) or six-week (2022) C2STEM curriculum as part of a program affiliated with Vanderbilt University. Each week, students attended a two-hour class where they received instruction on kinematics and computing from researchers at Vanderbilt University. Following the instructions, students completed various computational modeling tasks to reinforce their understanding.

During these modeling tasks, students' video and audio data were captured using OBS on laptops and lapel microphones. Their environment actions, captured as clickstream data, were logged using abstract syntax trees (ASTs) to provide a structural representation of their on-screen computational models. Each student action in the environment generated a corresponding change to the AST, with all actions recorded alongside timestamps, block IDs, and specific action types (e.g., adding, removing, editing). %
All students and parents provided informed assent and consent to participate. Vanderbilt University's IRB approved all study procedures. Demographic data such as race and gender could not be collected for these studies.

\subsection{Data Processing and SSRL Coding}

This paper focuses on integrating and analyzing students' environment logs and collaborative discourse. Raw discourse was automatically diarized and transcribed into utterance-level text using Otter.ai, then manually corrected by the authors of this paper. Raw actions were stored in ASTs and programmatically mapped to a cognitive task model adapted from prior work \cite{emara2021examining} to identify which of five low-level actions students performed:

\begin{itemize}
    \item \textbf{Build}: adding blocks to the computational model
    \item \textbf{Adjust}: moving, editing, or removing blocks from the computational model
    \item \textbf{Draft}: moving or editing blocks not connected to executable code (akin to ``commenting'' code)
    \item \textbf{Execute}: running (i.e., testing) the computational model
    \item \textbf{Visualize}: using the graph or table tools to inspect variable values
\end{itemize}

\begin{figure*}[htbp]
    \centering
    \includegraphics[width=\linewidth]{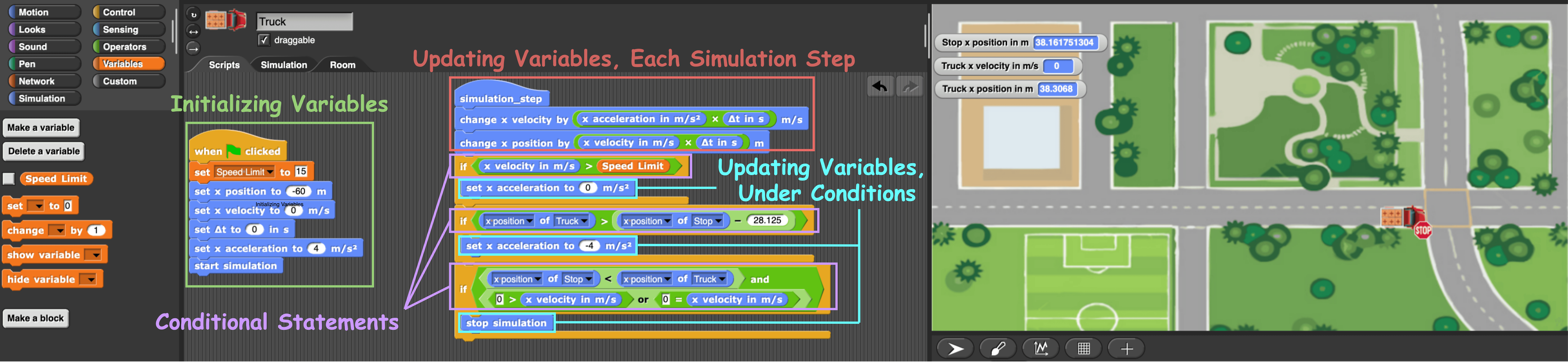}
    \caption{Example solution for the C2STEM Truck Task with Task Context categories \cite{cohn2025personalizing}.}
    \label{fig:c2stem}
    \Description[C2STEM learning environment.]{C2STEM learning environment. On the left are the blocks where students can drag various science and computing concepts (represented as code blocks) into the playground. In the middle is the playground, where students connect these blocks to build their computational models. On the right is a screen with a truck on a road, allowing students to visualize the truck's motion based on their model. The figure also indicates which blocks correspond to the Task Context categories. Any block connected to the ``When Green Flag Clicked'' block is treated as Initialization. The position and velocity update blocks directly under the Simulation Step block are treated as Updating Variables, Each Simulation Step. The ``if'' blocks beneath these, which handle conditions for the truck cruising, slowing down, and stopping, are treated as Conditional Statements. The blocks used to update the truck's variables (e.g., acceleration) inside the conditional statements are treated as Updating Variables, Under Conditions.}
\end{figure*}

Transcribed utterances and mapped actions were then aligned based on timestamps, and subsequently sampled for manual verification using screen and video recording data as ground truth. To group students' interactions temporally, we leveraged environment logs to segment each session according to \textit{Task Context} categories derived from prior work \cite{snyder2025using}
--- \textit{Initializing Variables; Updating Variables, Each Simulation Step; Updating Variables, Under Conditions; and Conditional Statements}---based on the specific portion of the computational model students were working on. Figure \ref{fig:c2stem} illustrates the C2STEM environment, an expert Truck Task solution, and these Task Context categories.

Each segment comprised a continuous sequence of actions within the same Task Context category, with a new segment beginning whenever students shifted their focus to a different part of the computational model. This approach enables contextualized segmentation, ensuring that each segment aligns with a consistent set of domain constructs and grounds student conversations in a shared problem-solving context, thereby mitigating the information loss often associated with traditional sliding window-based segmentation approaches that rely on arbitrary cut-offs.

For example, the "set x acceleration" block appears multiple times in the Truck Task computational model: once to initialize the acceleration variable and later within the simulation loop to update it depending on whether the truck is cruising or slowing down. The former requires understanding variable initialization, while the latter involves loops and conditional logic. Differentiating between these uses provides a richer context for interpreting both students' discourse and their environmental actions.

Once students' actions and utterances were integrated and segmented, we used GPT-3.5-Turbo---selected for its balance of cost and performance---to summarize each segment based on its discourse and environmental actions. The summarization prompt was adapted from prior work \cite{snyder2025using,snyder2024analyzing,white2023prompt} 
and is available for download in our supplementary materials\footnote{\href{https://github.com/claytoncohn/LAK26_Supplementary_Materials}{https://github.com/claytoncohn/LAK26\_Supplementary\_Materials}}. The prompt was dynamically adapted to each segment based on its task context derived from the environment logs, ensuring alignment between students' discourse and environment actions. %
All summaries were manually verified against the students' actions and discourse by members of the research team.

Once the LLM summaries were generated, we applied a previously defined coding scheme \cite{snyder2025using,snyder2024analyzing} to label them (see Table \ref{tab:coding_scheme}). The coding scheme is grounded in \cite{schunk1998self} and has been extended from analyzing individual SRL behaviors to examining group problem-solving, where students work in pairs to build complex computational models of scientific processes.  

To ensure agreement, two researchers assessed inter-rater reliability by randomly sampling 20 segment summaries and independently labeling them to calculate Cohen's $\kappa$. Discrepancies were resolved through discussion until consensus was reached. This process was repeated until Cohen's $\kappa \geq 0.7$, after which one researcher manually labeled the remaining segments. Each segment summary was assigned a single code from the coding scheme. The distributions of the seven codes range from 7\% to 22\%.

\begin{table}[htbp]
\caption{SSRL coding scheme \cite{snyder2025using}.}
\label{tab:coding_scheme}
\centering
    \begin{tabular}{p{0.27\linewidth}|p{0.67\linewidth}}    
    \textbf{Code} & \textbf{Description} \\
    \hline
    PLANNING  & Students decompose the problem and/or discuss steps to be taken to construct the simulation \\
    \hline
    ENACTING  & Students discuss the actions they are taking to construct the model \\
    \hline
    REFLECTING & Students review the simulation behavior and/or discuss the implications of their constructed code \\
    \hline
    ENACTING \& PLANNING   & Students plan and enact (construct) their solution in the same segment \\
    \hline
    ENACTING \& MONITORING & Students are building parts of the model and checking it in the same segment\\
    \hline
    ASSISTANCE & Students are receiving help from one of the researchers \\
    \hline
    OFF-TOPIC   & Students are having an off-topic discussion unrelated to the task  \\
    \hline
    \end{tabular}
\end{table}

Altogether, our study, data collection, and data processing yielded approximately 189 hours of collaborative discourse; 5,575 utterances; 4,893 logged actions (ADJUST=32\%, EXECUTE=32\%, BUILD=18\%, VISUALIZE=9\%, DRAFT=9\%); and 394 segments and summaries (\textit{Initializing Variables}=22\%; \textit{Updating Variables, Each Simulation Step}=17\%; \textit{Updating Variables, Under Conditions}=26\%; and \textit{Conditional Statements}=34\%) used for analysis. 

\begin{figure}[htbp]
    \centering
    \includegraphics[width=\linewidth]{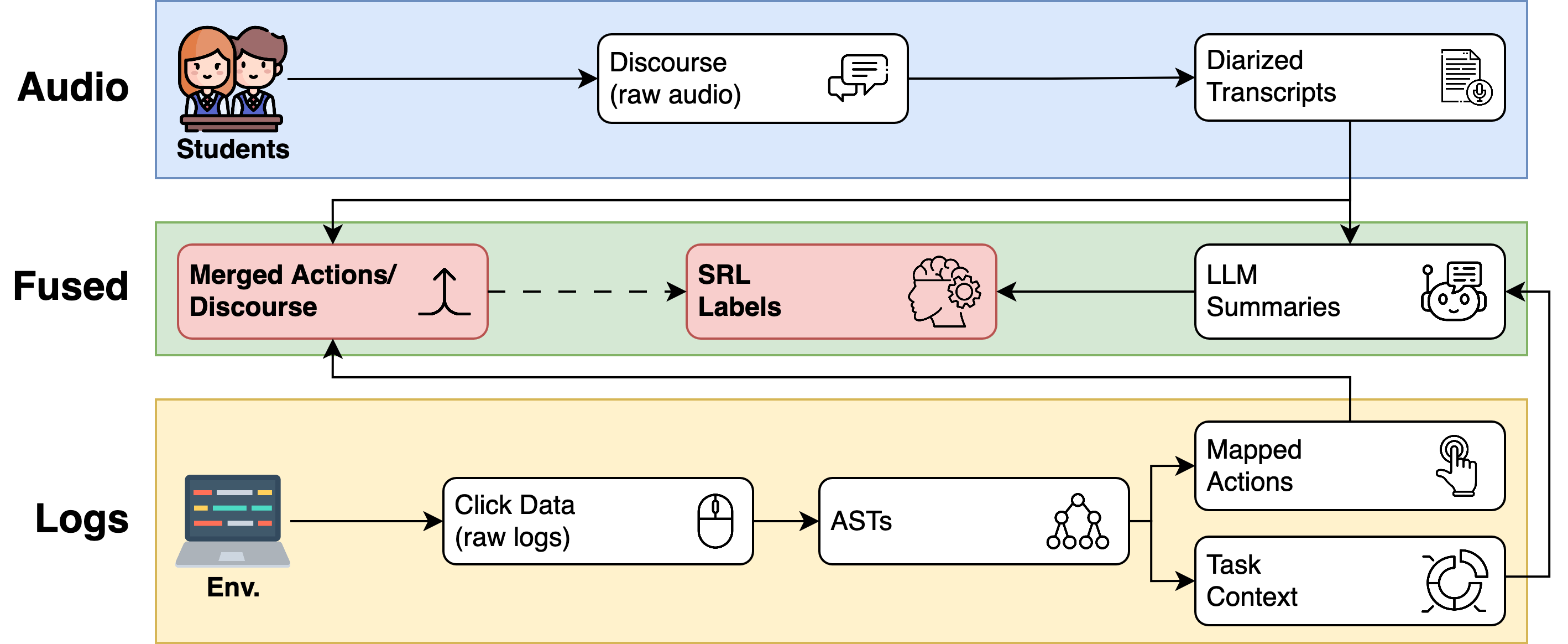}
    \caption{Multimodal data processing pipeline showing audio (blue) and logs (yellow) processing, and how the data was fused (green). Red boxes indicate data used for analysis in this paper (\textit{Merged Actions/Discourse} as inputs and \textit{SRL Labels} as outputs), which is represented by the dashed line and discussed shortly. All icons are c/o Flaticon.}
    \label{fig:pipeline}
    \Description[Multimodal data processing pipeline.]{Multimodal data processing pipeline. The pipeline comprises three vertically stacked components: audio processing (blue), environment logs processing (yellow), and fused data (green). The top component is audio. Students' collaborative discourse is recorded as raw audio files, which are then transcribed and diarized to produce text. The bottom component is logs. Students' clickstream data from the C2STEM environment is stored using ASTs. From these ASTs, we extract mapped actions corresponding to our cognitive task model, along with the Task Context categories used for segmentation. These categories indicate which part of the computational model students are working on at any given time. The middle component is fused data. First, mapped actions and discourse are time-aligned and merged. The diarized audio transcriptions and Task Context from the logs are then integrated into an LLM to generate contextualized summaries of each segment of student learning. These segment summaries are manually labeled using our coding scheme to identify the specific SSRL behaviors present in each segment. A dashed line connects the merged actions/discourse (inputs) to the SRL labels (outputs), representing the analysis conducted in this work.}
\end{figure}

Our multimodal pipeline employs \textit{mid fusion} \cite{cohn2024multimodal}, where input features are initially processed independently while remaining observable before being fused. This occurs twice in the pipeline: first, when mapped actions are time-aligned and integrated with collaborative discourse utterances; and second, when segment summaries are generated using the Task Context (derived from log data) alongside utterances. The complete data processing pipeline is shown in Figure \ref{fig:pipeline}.

\subsection{Predictive Modeling of SSRL Categories}

\subsubsection{Data Preparation}
Following best practices from prior work~\cite{borchers2024combining}, we constructed feature representations of collaborative dialogue segments for prediction. Utterances were embedded using the \verb|all-MiniLM-L6-v2| model from the \texttt{sentence-transformers} library, which maps text into dense numerical vectors that capture semantic similarity. To incorporate context, embeddings from the two preceding and two following segments were concatenated with the current segment, with padding as needed. In addition to embeddings, we engineered features from system log data, including sequential patterns of logged actions and uni-, bi-, and tri-gram representations of coded utterances aligned with the corresponding dialogue sequences. These log-derived features were designed to capture recurring procedural patterns that embeddings alone may not represent. Each segment was then merged with its annotated SSRL label, resulting in a dataset that supported comparisons across four representational conditions:

\textbf{(1) Text-only embeddings.} As a baseline, we trained models on the embedding of each utterance segment. This condition measured how well the semantic content of individual utterances predicted SSRL categories without contextual information.  

\textbf{(2) Context-enriched embeddings.} We concatenated embeddings from the two preceding and two following dialogue segments, thereby extending the representational space. This condition tested the hypothesis that semantic context improves predictive accuracy by capturing dependencies across sequential dialogue.  

\textbf{(3) Log-derived features.} We used features extracted from system log data and coded utterances, including uni-, bi-, and tri-gram patterns of interaction and their alignment with dialogue sequences. This condition allowed us to assess whether structural and behavioral traces of activity contribute additional predictive power beyond semantic content.  

\textbf{(4) Combined representations.} We constructed hybrid feature sets that integrated embeddings (with and without context) with log-derived features. This experiment evaluated whether combining semantic and structural information yields complementary benefits for prediction.  

The comparisons addressed two key questions: 1) Are text-only (semantic) embeddings alone sufficient for predicting SSRL categories, or does context substantially improve performance? 2) Does augmenting embeddings with log-derived features provide additional gains?
 
\subsubsection{Model Training and Evaluation}
Following best practices from prior work on SRL prediction \cite{zhang2024using}, we trained feed-forward neural networks implemented in \texttt{TensorFlow/Keras} to predict the presence and absence of each SSRL code. For each task, models featured two hidden layers (128--320 units), ReLU activations, dropout (0.1--0.3), and $L_{2}$ regularization. Optimization used the Adam optimizer with learning rates between $0.001$ and $0.01$, selected via randomized search. Models were trained for up to 100 epochs with early stopping based on validation AUC and adaptive learning-rate reduction.

To preserve evaluation integrity, we used leakage-safe, nested 3-fold cross-validation. In the outer loop, \texttt{GroupKFold} ensured all segments from the same collaborative session appeared only in either training or test sets. The inner loop performed randomized hyperparameter search, with imputation and standardization fit only on training folds, and class-balanced weights to address label imbalance. Batch sizes were adjusted between 8 and 32 for small or imbalanced folds. Model performance was evaluated using the area under the receiver operating characteristic curve (ROC AUC), given its robustness to class imbalance and interpretability in binary prediction contexts. Predictions from outer test folds were aggregated to compute distributions of ROC AUC, enabling systematic comparisons across feature configurations (text-only embeddings, context-enriched embeddings, log-derived features, and combinations).  

\section{Results}
\label{sec:results}

We evaluate five input configurations: (i) \textbf{text\_only} (utterance from the current segment), (ii) \textbf{text\_with\_context} (utterance with a $\pm 2$ segment window), (iii) \textbf{log\_only} (log features from the current segment), (iv) \textbf{log\_and\_text} (multimodal: utterance $+$ aligned log features from the current segment), and (v) \textbf{log\_and\_text\_context} (multimodal with a $\pm 2$ segment window). Models are trained and evaluated with nested cross-validation, and we report ROC~AUC with 95\% confidence intervals for each SSRL code (see Table \ref{tab:auc-by-code}).

\begin{table*}[htbp]
\centering
\caption{ROC AUC (95\% CI) by SSRL code and input configuration. The best performing models and configurations are highlighted in green.}
\label{tab:auc-by-code}
\resizebox{\textwidth}{!}{
\begin{tabular}{lcclcc}
\toprule
\textbf{SSRL Code} & \textbf{text\_only}& \textbf{text\_with\_context}&  \textbf{log\_only} &\textbf{log\_and\_text}& \textbf{log\_and\_text\_context}\\
\midrule
PLANNING & 0.5036 [0.3644, 0.6495] & \cellcolor{green!30}0.6711 [0.6058, 0.7230] &  0.4779 [0.3860, 0.5433] &0.4803 [0.4383, 0.5188] & 0.6269 [0.5538, 0.6944] \\
ENACTING & \cellcolor{green!30}0.6745 [0.6184, 0.7075] & 0.6113 [0.5364, 0.6538] &  0.5625 [0.5387, 0.5987] &0.6689 [0.6521, 0.6923] & 0.6102 [0.5940, 0.6389] \\
REFLECTING & 0.5158 [0.3954, 0.5880] & 0.5462 [0.4954, 0.6203] &  0.4735 [0.3246, 0.5559] &\cellcolor{green!30}0.5862 [0.4352, 0.8085] & 0.5795 [0.5117, 0.6242] \\
ENACT.\ \& PLAN. & \cellcolor{green!30}0.7285 [0.6855, 0.7584] & 0.6856 [0.5568, 0.7686] &  0.5898 [0.5613, 0.6293] &0.7240 [0.7016, 0.7362] & 0.6958 [0.6551, 0.7708] \\
ENACT.\ \& MONITOR. & \cellcolor{green!30}0.6796 [0.6435, 0.7192] & 0.5711 [0.4907, 0.6165] &  0.5753 [0.5443, 0.6289] &0.6501 [0.5555, 0.7271] & 0.6080 [0.5726, 0.6407] \\
ASSISTANCE & \cellcolor{green!30}0.6235 [0.5680, 0.6525] & 0.5895 [0.5094, 0.6391] &  0.5586 [0.4806, 0.6191] &0.5622 [0.4403, 0.6609] & 0.5639 [0.5180, 0.6102] \\
OFF-TOPIC & \cellcolor{green!30}0.8247 [0.7396, 0.9008] & 0.8090 [0.7493, 0.8645] &  0.4462 [0.2019, 0.6306] &0.7120 [0.6254, 0.8747] & 0.7782 [0.6239, 0.8912] \\
\bottomrule
\end{tabular}
}
\end{table*}

Across codes, aggregated mean AUCs (bootstrap 95\% CIs) are: \emph{text\_only} $=0.650$ [0.615, 0.685], \emph{text\_with\_context} $=0.640$ [0.598, 0.682], \emph{log\_only} $=0.526$ [0.471, 0.582], \emph{log\_and\_text} $=0.613$ [0.565, 0.661], and \emph{log\_and\_text\_context} $=0.623$ [0.579, 0.667]. These results show that \emph{text\_only} embeddings are strongest on average, but multimodal \emph{log\_and\_text} remains competitive for behaviors that are less likely to be verbally observed and for which environment interaction leaves clear traces (e.g., \emph{REFLECTING}: 0.5862). Text embeddings excel for enactment-oriented phases (\emph{ENACTING}: 0.6745; \emph{ENACTING \& PLANNING}: 0.7285; \emph{ENACTING \& MONITORING}: 0.6796), while short-range conversational context is particularly valuable for \emph{PLANNING} (utterance: 0.5036 $\rightarrow$ text\_with\_context: 0.6711; multimodal: 0.4803 $\rightarrow$ 0.6269). Notably, adding a $\pm2$ window to multimodal inputs can dilute temporally tight action-label coupling (e.g., \emph{ENACTING \& MONITORING}: 0.6501 $\rightarrow$ 0.6080). Overall, the detectors achieve practically useful accuracy for several SRL behaviors, with complementary strengths: multimodal traces (text$+$logs) are most discriminative when behaviors align with concrete environment use, and text embeddings—with carefully calibrated local context—capture conversationally grounded regulation phases.

\section{Discussion}
\label{sec:discussion}
This study advances learning analytics by demonstrating how embedding-based, multimodal representations can be leveraged to meaningfully detect socially shared regulation behaviors at scale in collaborative STEM+C learning. Below, we summarize our core findings, discuss their implications and limitations, and outline directions for future work.

\subsection{Main Findings}
In this study, we investigated the feasibility of training models to detect seven SSRL behaviors and group dynamics in a collaborative STEM+C learning environment \cite{cohn2024towards}. Using student discourse and corresponding log data, we sought to capture SSRL behaviors and processes, including planning, enacting, reflecting, enacting with planning, enacting with monitoring, providing assistance, and off-topic talk. To examine how different data sources contribute to model performance, we compared single-modality inputs (utterance-only and log-only) with multimodal inputs (utterance + log data). We further evaluated the role of contextual information by incorporating utterances from the two preceding and two following segments to enrich model representations. These methodological advances contribute to learning analytics by extending past methods for prediction of SRL (see \cite{zhang2024using}) to (a) open-ended, complex problem solving and (b) joint representations of log data and text artifacts produced by more than one student.

Our findings indicate that text-only models generally performed best across most SSRL categories, underscoring the central role of discourse in capturing how students communicate and regulate their learning together. Student utterances alone were often sufficient to distinguish key processes such as enacting or asking for assistance, suggesting that verbal exchanges provide a rich and reliable signal for modeling group regulation. At the same time, performance varied by construct (echoing findings from past research on multimodal prediction in our field \cite{wong2025rethinking}), pointing to the value of context and multimodality in specific cases. For example, planning behaviors benefited from incorporating surrounding utterances, which provided additional cues about students’ intentions and strategies. Reflective behaviors, in contrast, were better captured when log data were integrated, highlighting that certain forms of regulation may be enacted through interaction with the environment rather than explicitly verbalized. Together, these results suggest that while text-only approaches provide a strong and scalable baseline, the most effective strategy depends on the construct of interest. Contextual information and multimodal features can complement discourse analysis, offering deeper insight into behaviors that are less likely to be verbally articulated. 

Beyond accuracy, an important aspect of model fidelity in learning analytics is interpretability \cite{doshi2017towards}. As our multimodal models were often on par with text-only models, it is possible that their joint representation generates more useful insights for stakeholders such as teachers. Hence, it is important to understand how teachers perceive their usefulness and trustworthiness in practice. To this end, we gathered teacher feedback on both the usability of the models and the extent to which they would trust the outputs to inform instruction.

\subsection{Implications and Teacher Feedback}
To evaluate the practical utility of our approach, we engaged four teachers by presenting them, via email, with a complete segment of student data comprising student discourse, environment log actions, and predicted SSRL labels. The selected segment was intended to illustrate how discourse and log data can be time-aligned and integrated to provide a holistic view of student learning, and was presented to teachers as synchronized inputs corresponding to a specific SSRL label. 

Three of the participating teachers had over 20 years of teaching experience and more than five years of experience teaching computational modeling curricula using either the C2STEM kinematics curriculum or its Earth Science counterpart (including one teacher who contributed to the design of C2STEM), while the fourth teacher had five years of teaching experience and one year of experience teaching computational modeling with C2STEM. %
We posed the following guiding questions: 1) Do these behavior labels (\textit{Enacting} and \textit{Monitoring}) make sense in light of the provided discourse and actions? 2) If not, what additional information would support interpretation? 3) Does the integration of actions with discourse offer a more comprehensive view of student activity than either modality alone? 

Three of the four teachers affirmed that the student behaviors aligned with patterns they had observed in the discourse and environment log data. One teacher, however, expressed concern regarding the use of multiple labels within a single analytic segment. He argued this practice complicated interpretation by obscuring the behavioral signal in relation to the underlying data. Although prior research has demonstrated that SSRL behaviors frequently co-occur within the same learning episode \cite{snyder2024analyzing}, this critique highlights a critical challenge: such theoretical frameworks may not be immediately intuitive to practitioners. Future work must explore how to operationalize SSRL constructs in ways that are both interpretable and actionable for teachers, and investigate how learning analytics systems can be designed to better support practitioner needs in authentic classroom contexts.

Three teachers explicitly emphasized the value of integrating discourse data with environment log data to facilitate deeper sense-making and support more informed feedback to both teachers and students. As one teacher noted, \textit{``since both are connected...it does make sense to combine these actions when providing feedback.''} Another teacher elaborated that such integration \textit{``definitely provides a fuller picture than either modality alone''}, and suggested that the data could be even more actionable if presented visually --- such as through a teacher-facing dashboard or a multimodal timeline \cite{cohn2024multimodal}.

In contrast, one teacher focused almost exclusively on discourse, describing it as ``vital'' within the C2STEM environment. She observed that \textit{``students sometimes become rather isolated while coding''} and may become ``stuck,'' thus underscoring the pivotal role of discourse in supporting student progress and collaborative problem-solving. Collectively, these perspectives align with our findings: while discourse is central for identifying cognitive and metacognitive behaviors, the incorporation of environment log data offers insight into non-verbal actions. Our educational goal is not to classify student behaviors, but to design analytics that reveal meaningful patterns of student learning, enabling teachers to interpret these behaviors in context and translate them into timely, pedagogically actionable feedback. These modalities support a more holistic understanding of student behavior and enhance our ability to provide that feedback.

\subsection{Limitations and Future Work}
First, as the results indicate, different constructs benefit from different sources of information; some are captured well through discourse alone, while others rely more on contextual cues or multimodal data. Although the best-performing models achieved reasonable accuracy, there remains substantial room for improvement. As a next step, we propose exploring more advanced architectures (e.g., hierarchical models that better account for discourse flow) and richer multimodal features (e.g., gaze or gestures) to improve predictive performance. Second, we acknowledge limitations in the generalizability of the current study. The models were trained on data from students collaborating on a single task within a single learning environment. Future work should evaluate the transferability of this approach across subject domains, task types, and broader student populations to ensure its robustness in diverse classroom settings.

\section{Conclusion}
This work demonstrates the feasibility of using embedding-based deep learning to detect SSRL behaviors and group dynamics in collaborative STEM+C learning environments by integrating discourse and log data aligned with task context. By systematically comparing text-only, contextualized, and multimodal representations, we show that discourse provides a strong foundation for modeling, while context and log features offer complementary value for specific constructs. Beyond advancing scalable SSRL detection, our approach highlights the potential for multimodal learning analytics to provide educators with actionable insights into group regulation processes, ultimately supporting more effective feedback and intervention in collaborative classrooms.

\begin{acks}
This work is supported by the National Science Foundation under award DRL-2112635 and the Institute of Education Sciences through grant R305C240010. Any opinions, findings, and conclusions or recommendations expressed in this material are those of the authors and do not necessarily reflect the views of the National Science Foundation or the Institute of Education Sciences.
\end{acks}

\bibliographystyle{ACM-Reference-Format}
\bibliography{main}

\end{document}